\pdfoutput=1
\documentclass[11pt]{article}

\usepackage{emnlp2021}

\usepackage{times}
\usepackage{latexsym}

\usepackage[T1]{fontenc}
\usepackage[utf8]{inputenc}

\usepackage{microtype}

\usepackage{graphicx}
\usepackage{amsmath}
\usepackage{amsfonts}
\usepackage[linesnumbered,ruled,vlined]{algorithm2e}
\usepackage{multirow}
\usepackage{bbm}
\usepackage{tabularx}
\usepackage{enumitem}
\usepackage[british]{babel} 
\usepackage[misc]{ifsym}

\newcommand{\revised}[1]{{\color{green}{#1}}}

\title{ActiveEA: Active Learning for Neural Entity Alignment}

\author{Bing Liu\textsuperscript{1, \Letter}, Harrisen Scells\textsuperscript{1}, Guido Zuccon\textsuperscript{1}, Wen Hua\textsuperscript{1}, Genghong Zhao\textsuperscript{2} \\
\textsuperscript{1}The University of Queensland, Australia \\
\textsuperscript{2}Neusoft Research of Intelligent Healthcare Technology, Co. Ltd., China\\
\texttt{\{bing.liu, h.scells, g.zuccon, w.hua\}@uq.edu.au} \\
\texttt{zhaogenghong@neusoft.com}}

\begin{document}
\maketitle

\begin{abstract}
% background and motivation
Entity Alignment (EA) aims to match equivalent entities across different Knowledge Graphs (KGs) and is an essential step of KG fusion. Current mainstream methods -- neural EA models -- rely on training with seed alignment, i.e., a set of pre-aligned entity pairs which are very costly to annotate. %How to reduce the labour cost of annotating entities remains to study.
In this paper, we devise a novel Active Learning (AL) framework for neural EA, aiming to create highly informative seed alignment to obtain more effective EA models with less annotation cost. 
Our framework tackles two main challenges encountered when applying AL to EA: 

(1) How to exploit dependencies between entities within the AL strategy. Most AL strategies assume that the data instances to sample are independent and identically distributed. However, entities in KGs are related. To address this challenge, we propose a structure-aware uncertainty sampling strategy that can measure the uncertainty of each entity as well as its impact on its neighbour entities in the KG.

(2) How to recognise entities that appear in one KG but not in the other KG (i.e., \textit{bachelors}). Identifying bachelors would likely save annotation budget.
To address this challenge, we devise a bachelor recognizer paying attention to alleviate the effect of sampling bias.

Empirical results show that our proposed AL strategy can significantly improve sampling quality with good generality across different datasets, EA models and amount of bachelors.
\end{abstract}

\section{Introduction}

\begin{figure}
	\vspace{8pt}
    \centering
    \includegraphics[width=7.5cm]{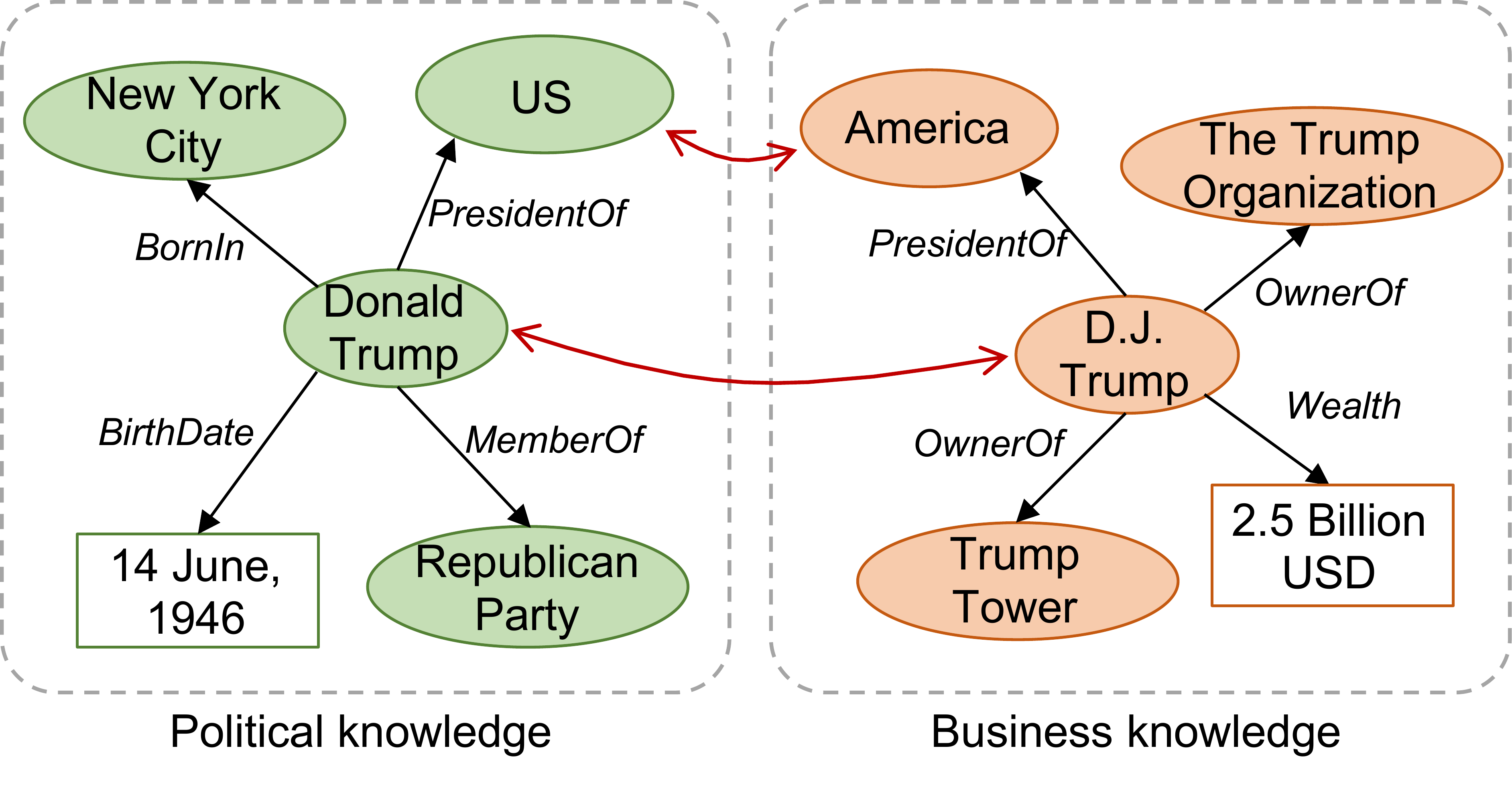}
    \vspace{-10pt}
    \caption{An example of Entity Alignment.\vspace{-8pt}}
    \label{fig:example}
\end{figure}

% Background 
Knowledge Graphs (KGs) store entities and their relationships with a graph structure and are used as knowledge drivers in many applications~\cite{DBLP:journals/corr/abs-2002-00388}.
Existing KGs are often incomplete but complementary to each other. 
A popular approach used to tackle this problem is KG fusion, which attempts to combine several KGs into a single, comprehensive one.
Entity Alignment (EA) is an essential step for KG fusion: it identifies equivalent entities across different KGs, supporting the unification of their complementary knowledge.
For example, in Fig.~\ref{fig:example} \textit{Donald Trump} and \textit{US} in the first KG correspond to \textit{D.J. Trump} and \textit{America} respectively in the second KG. By aligning them, the political and business knowledge about \textit{Donald Trump} can be integrated within one KG.

% For example, as shown in Fig.\ref{fig:example}, Wikipedia and IMDB both encode knowledge about the movie \textit{Titanic}. They share some basic information about the movie, e.g., the director and the main actors (thus, mates); but each KG also has content that is unique to it. Wikipedia contains entities associated with related stories like the inspiration and special effects of the movie, while IMDB contains entities related to rating and reviews from the audience (thus, bachelors). EA is used to link entities $A_1$ and $B_1$, $A_2$ and $B_2$, $A_3$ and $B_3$ so that the complementary knowledge about \textit{Titanic} can be integrated in a single KG.

% Existing works about EA and Motivation
%Neural EA methods~\cite{DBLP:conf/ijcai/ChenTYZ17,DBLP:conf/ijcai/ChenTCSZ18,DBLP:conf/emnlp/WangLLZ18,DBLP:conf/acl/CaoLLLLC19} have became the mainstream in recent years for their great potential of matching entities end-to-end.

Neural models~\cite{DBLP:conf/ijcai/ChenTYZ17,DBLP:conf/ijcai/ChenTCSZ18,DBLP:conf/emnlp/WangLLZ18,DBLP:conf/acl/CaoLLLLC19} are the current state-of-the-art in EA and are capable of matching entities in an end-to-end manner.
% that is, they do not rely on external attribute comparison tools like conventional methods~\cite{DBLP:conf/semweb/Jimenez-RuizG11,DBLP:journals/pvldb/SuchanekAS11}.
% Typically, neural EA methods are trained with an annotated seed alignment beforehand and then used to discover potential entity pairs. 
% However, the annotation process is a very labour-intensive task as it requires manual selection of entities from one KG and their equivalent counterparts in another KG.
Typically, these neural EA models rely on a seed alignment as training data which is very labour-intensive to annotate.
However, previous EA research has assumed the availability of such seed alignment and ignored the cost involved with their annotation.
%the cost issue of annotation has been neglected.
In this paper, we seek to reduce the cost of annotating seed alignment data, by investigating methods capable of selecting the most informative entities for labelling so as to obtain the best EA model with the least annotation cost: we do so using Active Learning.
% background about AL
Active Learning (AL)~\cite{DBLP:books/crc/aggarwal14/AggarwalKGHY14} is a Machine Learning (ML) paradigm where the annotation of data and the training of a model are performed iteratively so that the sampled data is highly informative for training the model.
Though many general AL strategies have been proposed~\cite{DBLP:series/synthesis/2012Settles,DBLP:journals/corr/abs-2009-00236}, there are some unique challenges in applying AL to EA.

% challenges
The first challenge is \textbf{how to exploit the dependencies between entities}. In the EA task, neighbouring entities (context) in the KGs naturally affect each other. For example, in the two KGs of Fig.~\ref{fig:example}, we can infer \textit{US} corresponds to \textit{America} if we already know that \textit{Donald Trump} and \textit{D.J. Trump} refer to the same person: this is because a single person can only be the president of one country. 
% a single person can only be the president of one country.
Therefore, when we estimate the value of annotating an entity, we should consider its impact on its context in the KG.
Most AL strategies assume data instances are independent, identically distributed and cannot capture dependencies between entities~\cite{DBLP:books/crc/aggarwal14/AggarwalKGHY14}. 
In addition, neural EA models exploit the structure of KGs in different and implicit ways~\cite{DBLP:journals/pvldb/SunZHWCAL20}. It is not easy to find a general way of measuring the effect of entities on others.

The second challenge is \textbf{how to recognize the entities in a KG that do not have a counterpart in the other KG} (i.e., \textit{bachelors}). 
In the first KG of Fig.~\ref{fig:example}, \textit{Donald Trump} and \textit{US} are matchable entities while \textit{New York City} and \textit{Republican Party} are bachelors. Selecting bachelors to annotate will not lead to any aligned entity pair.
The impacts of recognizing bachelors are twofold:

\begin{enumerate}[noitemsep,nolistsep,leftmargin=*]
	\item From the perspective of data annotation, recognizing bachelors would automatically save annotation budget (because annotators will try to seek a corresponding entity for some time before giving up) and allow annotators to put their effort in labelling matchable entities. This is particularly important for the existing neural EA models, which \textit{only} consider matchable entities for training: thus selecting bachelors in these cases is a waste of annotation budget.
	% Instead, they generate negative instances through negative sampling.% In this case, selecting bachelors to label would result in a waste of budget.
	%Even the strong assumption was dropped by works of the future, it is unnecessary to label them manually if we can identify them with model automatically.
	
	\item From the perspective of EA, bachelor recognition remedies the limitation of existing EA models that assume all entities to align are matchable, and would enable them to be better used in practice (i.e., real-life KGs where bachelors are popular).%(i.e., outside of artificial datasets where bachelors do not exist).
\end{enumerate}

% Annotating them will not produce any seed alignment. The existing neural EA models generate negative instances for their training process without considering using manually annotated bachelors. Therefore, annotating bachelors means a waste of budget. Meanwhile, bachelors are always the hard cases for models and thus have high probability of being selected under many general AL strategies. We claim bachelors bring significant side-effect and it would be beneficial if we can recognize and avoid selecting them. In addition, the assumption that every entity has a mate, which is based by existing neural EA models, is too strong in reality. Investigating bachelor identification would be a supplementary research to existing EA task.

% introduce our framework
To address these challenges, we propose a novel AL framework for EA. Our framework follows the typical AL process: entities are sampled iteratively, and in each iteration a batch of entities with the highest acquisition scores are selected. Our novel acquisition function consists of two components: a structure-aware uncertainty measurement module and a bachelor recognizer. 
The structure-aware uncertainty can reflect the uncertainty of a single entity as well as the influence of that entity in the context of the KG, i.e., how many uncertainties it can help its neighbours eliminate. 
% By combining the uncertainty sampling and the structure of the KG, our structure-aware uncertainty unifies their advantages in a single strategy. 
In addition, we design a bachelor recognizer, based on Graph Convolutional Networks (GCNs). Because the bachelor recognizer is trained with the sampled data and used to predict the remaining data, it may suffer from bias (w.r.t. the preference of sampling strategy) of these two groups of data. We apply model ensembling to alleviate this problem.

Our major contributions in this paper are:
\begin{enumerate}[noitemsep,nolistsep,leftmargin=*]
\item A novel AL framework for neural EA, which can produce more informative data for training EA models while reducing the labour cost involved in annotation. To our knowledge, this is the first AL framework for neural EA.
\item A structure-aware uncertainty sampling strategy, which models uncertainty sampling and the relation between entities in a single AL strategy.
\item An investigation of bachelor recognition, which can reduce the cost of data annotation and remedy the defect of existing EA models.
\item Extensive experimental results that show our proposed AL strategy can significantly improve the quality of data sampling and has good generality across different datasets, EA models, and bachelor quantities.
\end{enumerate}

\section{Background}

\vspace{-6pt}
\subsection{Entity Alignment}\label{sec:preliminary_ea}

Entity alignment is typically performed between two KGs $\mathcal{G}^1$ and $\mathcal{G}^2$, whose entity sets are denoted as $\mathcal{E}^1$ and $\mathcal{E}^2$ respectively. The goal of EA is to find the equivalent entity pairs $\mathcal{A} = \{(e^1, e^2) \in \mathcal{E}^1 \times \mathcal{E}^2 | e^1 \sim e^2\}$, where $\sim$ denotes an equivalence relationship and is usually assumed to be a one-to-one mapping. In supervised and semi-supervised models, a subset of the alignment $\mathcal{A}^{seed} \subset \mathcal{A}$, called seed alignment, are annotated manually beforehand and used as training data. The remaining alignment form the test set $\mathcal{A}^{test} = \mathcal{A} \setminus \mathcal{A}^{seed}$.
The core of an EA model $F$ is a scoring function $F(e^1, e^2)$, which takes two entities as input and returns a score for how likely they match.
The effectiveness of an EA model is essentially determined by $\mathcal{A}^{seed}$ and we thus denote it as $m(\mathcal{A}^{seed})$.

\vspace{-4pt}
\subsection{Active Learning}
% Active Learning (AL) is a general framework of data selection, which aims to select training data for given a model $F$ dynamically, effectively and efficiently.
An AL framework consists of two components: (1) an \textit{oracle} (annotation expert), which provides labels for the \textit{queries} (data instances to label), and (2) a \textit{query} system, which selects the most informative data instances as queries. 
In pool-based scenario, there is a pool of unlabelled data $\mathcal{U}$. Given a budget $B$, some instances $\mathcal{U}_{\pi,B}$ are selected from the pool following a strategy $\pi$ and sent to the experts to annotate, who produce a training set  $\mathcal{L}_{\pi,B}$. 
We train the model on $\mathcal{L}_{\pi,B}$ and the effectiveness $m(\mathcal{L}_{\pi,B})$ of the obtained model reflects how good the strategy $\pi$ is. The goal is to design an optimal strategy $\pi_*$ such that $ \pi_* = \mathrm{argmax}_{\pi} m(\mathcal{L}_{\pi, B}) \text{.}$

% In general, AL is an iterative process. In each iteration, 1) the strategy, which might be based on the prediction results of the current model, is applied to estimate the informativeness of each data instance and a certain number of instances with the highest informativeness are selected for labelling; 2) the newly selected data are labelled and added to the training set; 3) the strategy is upgraded, for example by updating the model with the newest training data, and the process enters the next interation till all budgets are consumed. 

% The \textit{acquisition function}, which is used to estimate the value of each data instance, lies in the center of AL strategies. 

% For example, \text{uncertanity sampling}, one of the most popular strategy, uses the uncertainty of current model on each data instance as its acquisition function.

\begin{figure}
    \centering
    \includegraphics[width=7.5cm]{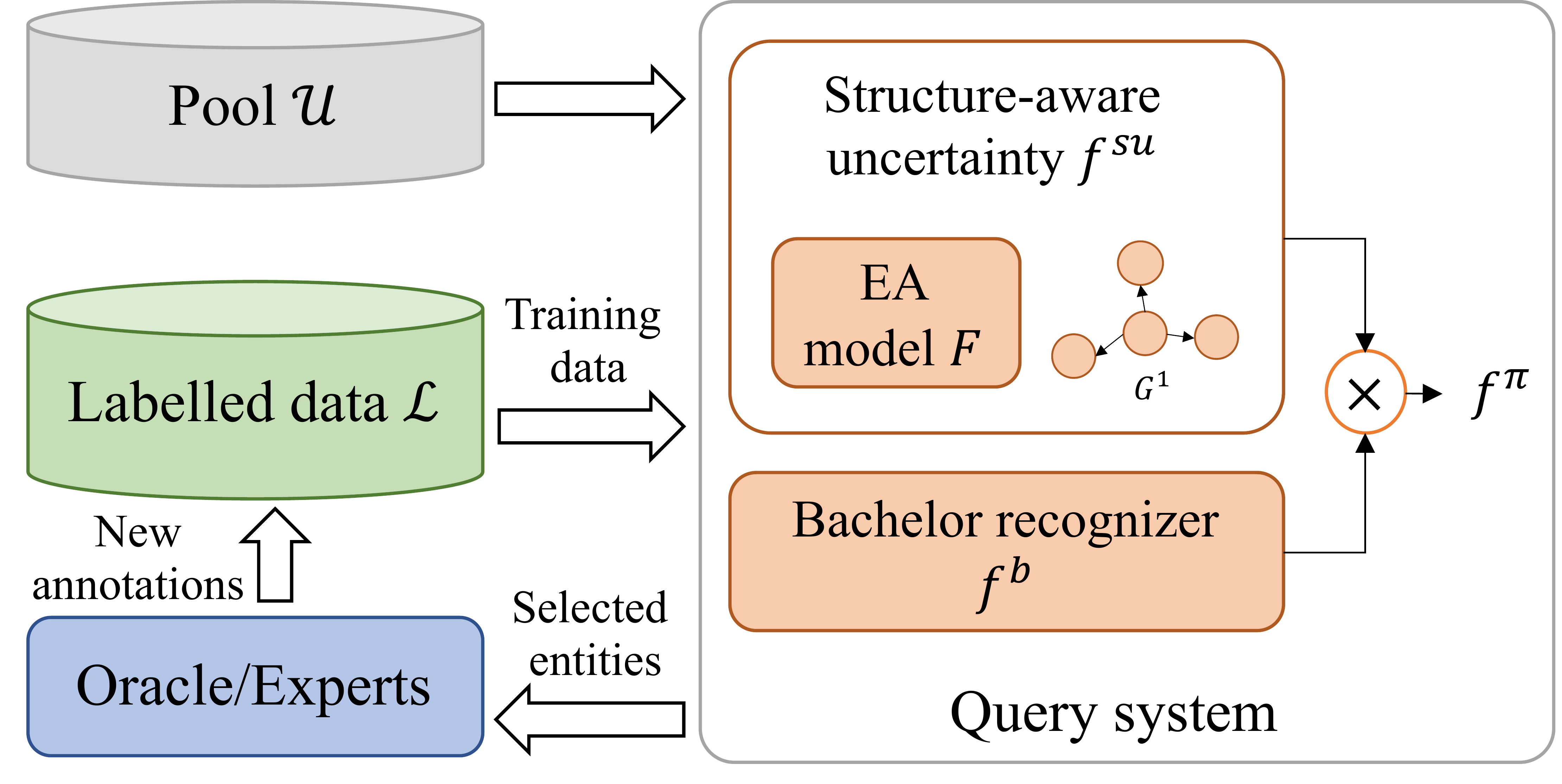}
    \vspace{-8pt}
    \caption{Overview of ActiveEA.\vspace{-16pt}}
    \label{fig:overview}
\end{figure}

\section{ActiveEA: Active Entity Alignment}
\vspace{-4pt}
\subsection{Problem Definition}

Given two KGs $\mathcal{G}^1$, $\mathcal{G}^2$ with entity sets $\mathcal{E}^1$, $\mathcal{E}^2$, an EA model $F$, a budget $B$, the AL strategy $\pi$ is applied to select a set of entities $\mathcal{U}_{\pi, B}$ so that the annotators label the counterpart entities to obtain the labelled data $\mathcal{L}_{\pi, B}$.
% The pairs of entities from $\mathcal{G}^1$, $\mathcal{G}^2$ that can be matched form a seed alignment $\mathcal{A}^{seed}_{\pi,B}$.
$\mathcal{L}_{\pi, B}$ consists of annotations of matchable entities $\mathcal{L}^+_{\pi, B}$, which form the seed alignment $\mathcal{A}^{seed}_{\pi,B}$, and bachelors $\mathcal{L}^-_{\pi, B}$.
We measure the effectiveness  $m(\mathcal{A}^{seed}_{\pi, B})$ of the AL strategy $\pi$ by training the EA model on $\mathcal{A}^{seed}_{\pi, B}$ and then evaluating it with $\mathcal{A}^{test}_{\pi,B}=\mathcal{A} \setminus \mathcal{A}^{seed}_{\pi,B}$.
Our goal is to design an optimal entity sampling strategy $\pi_*$ so that $\pi_*=\mathrm{argmax}_{\pi} m(\mathcal{A}^{seed}_{\pi,B})$.

In our annotation setting, we select entities from one KG and then let the annotators identify their counterparts from the other KG. 
% We assume the annotators are equipped with the ability of high-quality annotation and the cost of labelling each entity is fixed.
Under this setting, we assume the pool of unlabelled entities is initialized with $\mathcal{U}=\mathcal{E}^1$.
The labelled data will be like $\mathcal{L}^+_{\pi, B}=\{ (e^1 \in \mathcal{E}^1, e^2 \in \mathcal{E}^2 ) \}$ and $\mathcal{L}^-_{\pi, B}=\{ (e^1 \in \mathcal{E}^1, null ) \}$.
% In each sampling iteration, the sampling batch size is $N$.

% The pool usually contains both mate entities and bachelor entities, which are denoted as $\mathcal{P}^+$ and $\mathcal{P}^-$ respectively.

\vspace{-4pt}
\subsection{Framework Overview}
The whole annotation process, as shown in Fig.~\ref{fig:overview}, is carried out iteratively. In each iteration, the query system selects $N$ entities from $\mathcal{U}$ and sends them to the annotators. The query system includes (1) a structure-aware uncertainty measurement module $f^{su}$, which combines uncertainty sampling with the structure information of the KGs, and (2) a bachelor recognizer $f^b$, which helps avoid selecting bachelor entities. The final acquisition $f^{\pi}$ used to select which entities to annotate is obtained by combining the outputs of these two modules. After the annotators assign the ground-truth counterparts to the selected entities, the new annotations are added to the labelled data $\mathcal{L}$. With the updated $\mathcal{L}$, the query system updates the EA model and the bachelor recognizer. This process repeats until no budget remains. 
To simplify the presentation, we omit the sampling iteration when explaining the details.% of our method. 

\subsection{Structure-aware Uncertainty Sampling}\label{sec:structure_uncertainty}

We define the influence of an entity on its context as the amount of uncertainties it can help its neighbours remove. 
As such, we formulate the structure-aware uncertainty $f^{su}$ as

    \vspace{-12pt}
\begin{equation}
    \begin{split}
        f^{su}(e^1_i) & =  \alpha \sum_{e^1_i \rightarrow e^1_j, e^1_j \in \mathcal{N}^{out}_{i}} w_{ij} f^{su}(e^1_j) \\ 
        & + (1-\alpha) \frac{f^u(e^1_i)}{\sum_{e^1 \in \mathcal{E}^1} f^u(e^1)} \text{,}
    \end{split}
    \label{eq:pagerank}
\end{equation}
where $\mathcal{N}^{out}_i$ is the outbound neighbours of entity $e^1_i$ (i.e. the entities referred to by $e^1_i$) and $w_{ij}$ measures the extent to which $e^1_i$ can help $e^1_j$ eliminate uncertainty. 
The parameter $\alpha$ controls the trade-off between the impact of entity $e^1_i$ on its context (first term in the equation) and the normalized uncertainty (second item). Function $f^u(e^1)$ refers to the margin-based uncertainty of an entity. For each entity $e^1$, the EA model can return the matching scores $F(e^1, e^2)$ with all unaligned entities $e^2$ in $\mathcal{G}^2$. Since these scores in existing works are not probabilities, we exploit the margin-based uncertainty measure for convenience, outlined in Eq.~\ref{eq:uncertainty}: %gives a formal definition of our margin-based uncertainty measure:
\vspace{-8pt}
\begin{equation}
    f^u(e^1) = - \left(F(e^1, e^2_*) - F(e^1, e^2_{**}) \right)
    \label{eq:uncertainty}
\end{equation}
where $F(e^1, e^2_*)$ and $F(e^1, e^2_{**})$ are the highest and second highest matching scores respectively. A large margin represents a small uncertainty.

For each entity $e^1_j$, we assume its inbound neighbours can help it clear all uncertainty. Then, we have 
$ \sum_{e^1_i \rightarrow e^1_j, e^1_i \in \mathcal{N}^{in}_j} w_{ij} = 1 \text{,}$
where $\mathcal{N}^{in}_j$ is the inbound neighbour set of $e^1_j$.
In this work, we assume all inbound neighbours have the same impact on $e^1_j$. In this case,
$ w_{ij} = \frac{1}{\mathrm{degree}(e^1_j)}$, where $\mathrm{degree}(\cdot)$ returns the in-degree of an entity.

Using matrix notion, Eq.~\ref{eq:pagerank} can be rewritten as
\vspace{-8pt}
\begin{equation*}
    \mathbf{f}^{su} = \alpha \mathbf{W} \mathbf{f}^{su} + (1-\alpha) \frac{\mathbf{f}^{u}}{|\mathbf{f}^{u}|} 
%    \label{eq:pagerank_matrix}
\vspace{-4pt}
\end{equation*}
where $\mathbf{f}^{su}$ is the vector of structure-aware uncertainties, $\mathbf{f}^{u}$ is the vector of uncertainties, and $\mathbf{W}$ is a matrix encoding influence between entities, i.e., $w_{ij}>0$ if $e^1_i$ is linked to $e^1_j$, otherwise 0.

As $\mathbf{W}$ is a stochastic matrix~\cite{gagniuc2017markov}, we solve Eq.~\ref{eq:pagerank} iteratively, which can be viewed as the power iteration method~\cite{DBLP:journals/cacm/Franceschet11}, similar to \textit{Pagerank}~\cite{DBLP:journals/cn/BrinP98}.
Specifically, we initialize the structure-aware uncertainty vector as $\mathbf{f}_0^{su}=\mathbf{f}^u$. Then we update $\mathbf{f}_{t}^{su}$ iteratively:
\begin{equation*}
    \mathbf{f}_{t}^{su} =  \alpha \mathbf{W} \mathbf{f}_{t-1}^{su} + (1-\alpha) \frac{\mathbf{f}^{u}}{|\mathbf{f}^{u}|} , t=1,2,3,...
%    \label{eq:iterative_pagerank}
\end{equation*}
The computation ends when $|\mathbf{f}_{t}^{su} - \mathbf{f}_{t-1}^{su}| < \epsilon$.% for a small $\epsilon$.

\subsection{Bachelor Recognizer}
The bachelor recognizer is formulated as a binary classifier, which is trained with the labelled data and used to predict the unlabelled data.
One challenge faced here is the bias between the labelled data and the unlabelled data caused by the sampling strategy (since it is not random sampling). We alleviate this issue with a model ensemble.

\subsubsection{Model Structure}

We apply two GCNs~\cite{DBLP:conf/iclr/KipfW17,DBLP:conf/nips/HamiltonYL17}  as the encoders to get the entity embeddings $\mathbf{H}^1 = \mathbf{GCN}^1(\mathcal{G}^1), \mathbf{H}^2 = \mathbf{GCN}^2(\mathcal{G}^2)$, where each row in $\mathbf{H}^1$ or $\mathbf{H}^2$ corresponds to a vector representation of a particular entity. 
The two GCN encoders share the same structure but have separate parameters. With each GCN encoder, each entity $e_i$ is first assigned a vector representation $\mathbf{h}^{(0)}_i$. Then contextual features of each entity are extracted:
\begin{equation*}
    \mathbf{h}^{(l)}_i = \mathrm{norm}( \sigma ( \sum_{j \in \mathcal{N}_i \cup \{i\}} \mathbf{V}^{(l)} \mathbf{h}^{(l-1)}_j + \mathbf{b}^{(l)} ) ) \textit{,}
\end{equation*}
where $l$ is the layer index, $\mathcal{N}_i$ is the neighbouring entities of entity $e_i$, and $\sigma$ is the activation function, $\mathrm{norm}(\cdot)$ is a normalization function, and $\mathbf{V}^{(l)}, \mathbf{b}^{(l)}$ are the parameters in the $l$-th layer. The representations of each entity $e_i$ obtained in all GCN layers are concatenated into a single representation: $\mathbf{h}_i = \mathrm{concat}(\mathbf{h}^{(0)}_i, \mathbf{h}^{(1)}_i, ..., \mathbf{h}^{(L)}_i)$, where $L$ is the number of GCN layers.

After getting the representations of entities, we compute the similarities of each entity in $\mathcal{E}^1$ with all entities in $\mathcal{E}^2$ ($\mathbf{S} = \mathbf{H}^1 \cdot {\mathbf{H}^2}^T$) and obtain its corresponding maximum matching score as in $f^s(e^1_i) = \max (\mathbf{S}_{i, :})$. The entity $e_i^1$ whose maximum matching score is greater than a threshold $\gamma$ is considered to be a matchable entity as in $f^b(e^1_i) = \mathbbm{1}_{f^s(e^1_i) > \gamma}$, otherwise a bachelor.

\subsubsection{Learning}
In each sampling iteration, we train the bachelor recognizer with existing annotated data $\mathcal{L}$ containing  matchable entities $\mathcal{L}^+$ and bachelors $\mathcal{L}^-$.
% $\mathcal{L}^+=\{ (e^1 \in \mathcal{E}^1, e^2 \in \mathcal{E}^2) \}$ and bachelors $\mathcal{L}^-=\{ (e^1 \in \mathcal{E}^1, null) \}$.
Furthermore, $\mathcal{L}$ is divided into a training set $\mathcal{L}^t$ and a validation set $\mathcal{L}^v$. 

We optimize the parameters, including $\{ \mathbf{V}^{(l)}, \mathbf{b}^{(l)} \}_{1 \leq l \leq L}$ of each GCN encoder and the threshold $\gamma$, in two phases, sharing similar idea with supervised contrastive learning~\cite{DBLP:conf/nips/KhoslaTWSTIMLK20}.
In the first phase, we optimize the scoring function $f^s$ by minimizing the constrastive loss  shown in Eq.~\ref{eq:loss_of_bachelor_recognizer}. %with the seed alignment in the training set $\mathcal{L}^{t, +}$. 

\vspace{-14pt}
\begin{equation}
    \begin{split}
        loss & = \sum_{(e^1_i, e^2_j) \in \mathcal{L}^{t, +} } \parallel \mathbf{h}^1_i - \mathbf{h}^2_j \parallel \\
        & + \beta \sum_{(e^1_{i'}, e^2_{j'}) \in \mathcal{L}^{t, neg}} [\lambda - \parallel \mathbf{h}^1_{i'} - \mathbf{h}^2_{j'} \parallel ]_+
    \end{split}
    \vspace{-12pt}
    \label{eq:loss_of_bachelor_recognizer}
\end{equation}
Here, $\beta$ is a balance factor, and $[\cdot]_+$ is $\max(0, \cdot)$, and $\mathcal{L}^{t, neg}$ is the set of negative samples generated by negative sampling~\cite{DBLP:conf/ijcai/SunHZQ18}.
For a given pre-aligned entity pair in $\mathcal{L}^+$, each entity of it is substituted for $N^{neg}$ times.
The distance of negative samples is expected to be larger than the margin $\lambda$. 
In the second phase, we freeze the trained $f^s$ and optimize $\gamma$ for $f^b$.
It is easy to optimize $\gamma$, e.g. by simple grid search, so that $f^b$ can achieve the highest performance on $\mathcal{L}^v$ (denoted as $q(f^s, \gamma, \mathcal{L}^v)$) using:
\vspace{-8pt}
\begin{equation*}
    \gamma^* = \mathrm{argmax}_{\gamma} q(f^s, \gamma, \mathcal{L}^v) \text{.}
\end{equation*}

\subsubsection{Model Ensemble for Sampling Bias}
The sampled data may be biased, since they have been preferred by the sampling strategy rather than selected randomly.
As a result, even if the bachelor recognizer is well trained with the sampled data it may perform poorly on data yet to sample.
We apply a model ensemble to alleviate this problem. 
Specifically, we divide the $\mathcal{L}$ into $K$ subsets evenly. Then we apply $K$-fold cross-validation to train $K$ scoring functions $\{ f^s_1, ..., f^s_K \}$, each time using $K-1$ subsets as the training set and the left out portion as validation set. 
Afterwards, we search for an effective $\gamma$ threshold:
\vspace{-8pt}
\begin{equation*}
    \gamma^* = \mathrm{argmax}_{\gamma} \frac{1}{K} \sum_{1 \leq k \leq K} q(f^s_k, \gamma, \mathcal{L}^v_k)
    \vspace{-10pt}
\end{equation*}

%In the inference stage, we ensemble the $n$ scoring functions $f^s_i$ by averaging them to form the final scoring function and base $f^b$ on it:

At inference, we ensemble by averaging the $K$ scoring functions $f^s_k$ to form the final scoring function $f^s$ as in Eq.~\ref{eq:ave_bachelor_scoring} and base $f^b$ on it.

\vspace{-8pt}
\begin{equation}
    f^s(e^1_i) = \frac{1}{K} \sum_{1 \leq k \leq K} f^s_k (e^1_i)
   \label{eq:ave_bachelor_scoring}
\vspace{-4pt}
\end{equation}

\subsection{Final Acquisition Function}  % add a section title to address our final acquisition function is a combination of the above two modules

We combine our structure-aware uncertainty sampling with the bachelor recognizer to form the final acquisition function:
\vspace{-8pt}
$$
    f^{\pi}(e^1_i) = f^{su}(e^1_i) f^b (e^1_i) 
     \label{eq:acquisition_function}
$$
\vspace{-14pt}

\section{Experimental Setup}
%We established some sampling strategies and their running settings, which were different combinations of EA models, datasets, and bachelor percentages.

\subsection{Sampling Strategies}

We construct several baselines for comparison:
\begin{description}[noitemsep,nolistsep,leftmargin=*]
\item[\textit{rand}] random sampling used by existing EA works.
\item[\textit{degree}] selects entities with high degrees.
\item[\textit{pagerank}]~\cite{DBLP:journals/cn/BrinP98} measures the centrality of entities by considering their degrees as well as the importance of its neighbours.
\item[\textit{betweenness}]~\cite{freeman1977set} refers to the number of shortest paths passing through an entity.
\item[\textit{uncertainty}] sampling selects entities that the current EA model cannot predict with confidence. Note that in this work we measure uncertainty using Eq.~\ref{eq:uncertainty} for fair comparison.
\end{description}

\textit{degree}, \textit{pagerank} and \textit{betweenness} are purely topology-based and do not consider the current EA model. On the contrary, \textit{uncertainty} is fully based on the current EA model without being able to capture the structure information of KG.
We compare both our structure-aware uncertainty sampling (\textbf{\textit{struct\_uncert}}) and the full framework \textbf{\textit{ActiveEA}} with the baselines listed above.   
We also examine the effect of \textbf{\textit{Bayesian Transformation}}, which aims to make deep neural models represent uncertainty more accurately~\cite{DBLP:conf/icml/GalIG17}. %, which was proposed to make the uncertainty of deep neural models more precise.
% ~\cite{DBLP:conf/www/OstapukYC19}

%We also examine the effect of \textbf{\textit{Bayesian transformation}}, which aims to make deep neural models produce more accurate confidence estimations~\cite{DBLP:conf/icml/GalIG17}. %, which was proposed to make the uncertainty of deep neural models more precise.

\subsection{EA Models}

We apply our ActiveEA framework to three different EA models, which are a representative spread of neural EA models and varied in KG encoding, considered information and training method~\cite{DBLP:conf/emnlp/LiuCPLC20,DBLP:conf/ijcai/SunHZQ18}:
% Table \ref{tab:selected_ea_models} summarizes their characteristics w.r.t different aspects.
\begin{description}[noitemsep,nolistsep,leftmargin=*]
	\item[BootEA]~\cite{DBLP:conf/ijcai/SunHZQ18} encodes the KGs with the translation model~\cite{DBLP:conf/nips/BordesUGWY13}, exploits the structure of KGs, and uses self-training.
	\item[Alinet]~\cite{DBLP:conf/aaai/SunW0CDZQ20} also exploits the structure of KGs but with a GCN-based KG encoder, and is trained in a supervised manner.
	\item[RDGCN]~\cite{DBLP:conf/ijcai/WuLF0Y019} trains a GCN in a supervised manner, as Alinet, but it can incorporate entities' attributes.
\end{description}
% BootEA models the KG structure using a translation \todo{language(?)} model~\cite{DBLP:conf/nips/BordesUGWY13}, a traditional technique. Meanwhile, RDGCN and Alinet apply more contemporary GNN-based models~\cite{DBLP:conf/iclr/KipfW17}.
% BootEA and Alinet only consider the structure of KG, while RDGCN exploits the attributes of entities.
% RDGCN and Alinet only use the provided seed alignments as training data, while BootEA augments the seeds using semi-supervised method.
%These models' 
Our implementations and parameter settings of the models rely on  OpenEA\footnote{\url{https://github.com/nju-websoft/OpenEA}}~\cite{DBLP:journals/pvldb/SunZHWCAL20}.

\subsection{Datasets}

We use three different datasets: D-W-15K V1 (\textit{DW}), EN-DE-15K V1 (\textit{ENDE}), and EN-FR-100K V1 (\textit{ENFR}), obtained from OpenEA~\cite{DBLP:journals/pvldb/SunZHWCAL20}.
%They are denoted as \textit{DW}, \textit{ENDE}, and \textit{ENFR} respectively in this work.
Each dataset contains two KGs and equivalent entity pairs.
The KGs used in these datasets were sampled from real KGs, i.e. \textit{DBpedia}~\cite{DBLP:journals/semweb/LehmannIJJKMHMK15}, \textit{Wikidata}~\cite{DBLP:journals/cacm/VrandecicK14}, and \textit{YAGO}~\cite{DBLP:conf/semweb/RebeleSHBKW16}, which are widely used in EA community.
These datasets differ in terms of KG sources, languages, sizes, etc. We refer the reader to \citet{DBLP:journals/pvldb/SunZHWCAL20} for more details.

% \textit{DW} is a monolingual EA dataset aligning \textit{DBpedia} to \textit{Wikidata}, while \textit{ENDE} and \textit{ENFR} are cross-lingual datasets which aligned the English version of \textit{DBpedia} to its German version and French version respectively.
% Table~\ref{tab:statistics_of_datasets} shows their more statistics in terms of the number of relations, attributes, triples, etc.

Existing work on EA assumes all entities in the KGs are matchable, thus only sampling entities with counterparts when producing the datasets.
For investigating the influence of bachelors on AL strategies, we synthetically modify the datasets by excluding a portion of entities from the second KG.% which results in bachelors in the first KG.
% \todo{TODO: more arguments for this method of generating bachelors}

% \begin{table}[tb]
%     \caption{Characteristics of Selected EA models}
%     \begin{tabular}{|c|c|c|c|c|}
%         \hline
%         Name  & KG modelling & Structure & Attribute & Semi \\
%         \hline
%         BootEA & Trans & $\checkmark$ & & $\checkmark$ \\
%         \hline
%         Alinet  & GNN & $\checkmark$ & & \\
%         \hline
%         RDGCN & GNN & $\checkmark$ & $\checkmark$ & \\
%         \hline
%         AttrGNN & GNN & $\checkmark$ & $\checkmark$ & \\
%         \hline
%     \end{tabular}
%     \label{tab:selected_ea_models}
% \end{table}

% \begin{table}[tb]
%     \caption{Dataset Statistics.}

%     \begin{tabular}{|cc|rrrrr|}
%         \hline
%              \multicolumn{2}{|c|}{Datasets} & \#Ent. & \#Rel. & \#Att. & \#Rel tr. & \#Att tr. \\ 
%         \hline 
%         \multirow{2}{*}{DW} & $\mathcal{G}^1$ & 15K & 248 & 342 & 38,265 & 68,258  \\
%             & $\mathcal{G}^2$ & 15K & 169 & 649 & 42,746 & 138,246  \\
%         \hline
%         \multirow{2}{*}{ENDE}  & $\mathcal{G}^1$ & 15K & 215 & 286 & 47,676 & 83,755   \\
%             & $\mathcal{G}^2$ & 15K & 131 & 194 & 50,419 & 156,150   \\
%         \hline 
%         \multirow{2}{*}{ENFR} & $\mathcal{G}^1$ & 100K & 400 & 466 & 309,607 & 497,729  \\
%             & $\mathcal{G}^2$ & 100K & 300 & 519 & 258,285 & 426,672  \\
%         \hline

%     \end{tabular}

%     \label{tab:statistics_of_datasets}
% \end{table}

\subsection{Evaluation Metrics}

We use Hit@1 as the primary evaluation measure of the EA models. To get an overall evaluation of one AL strategy across different sized budgets, we plot the curve of a EA model's effectiveness with respect to the proportion of annotated entities, and calculate the Area Under the Curve (AUC).% at some concerned proportions.  We compare different strategies with their curves and AUC values.

% \todo{TODO: signficance test}

\subsection{Parameter Settings}

We set $\alpha=0.1$, $\epsilon=1e^{-6}$ for the structure-aware uncertainty.
We use $L=1$ GCN layer for our bachelor recognizer with 500 input and 400 output dimensions. We set $K=5$ for its model ensemble and $\lambda=1.5$, $\beta=0.1$, $N^{neg}=10$ for its training. 
The sampling batch size is set to $N=100$ for 15K data and $N=1000$ for 100K data.

\begin{figure*}[t!]
	\centering
	\includegraphics[width=16cm]{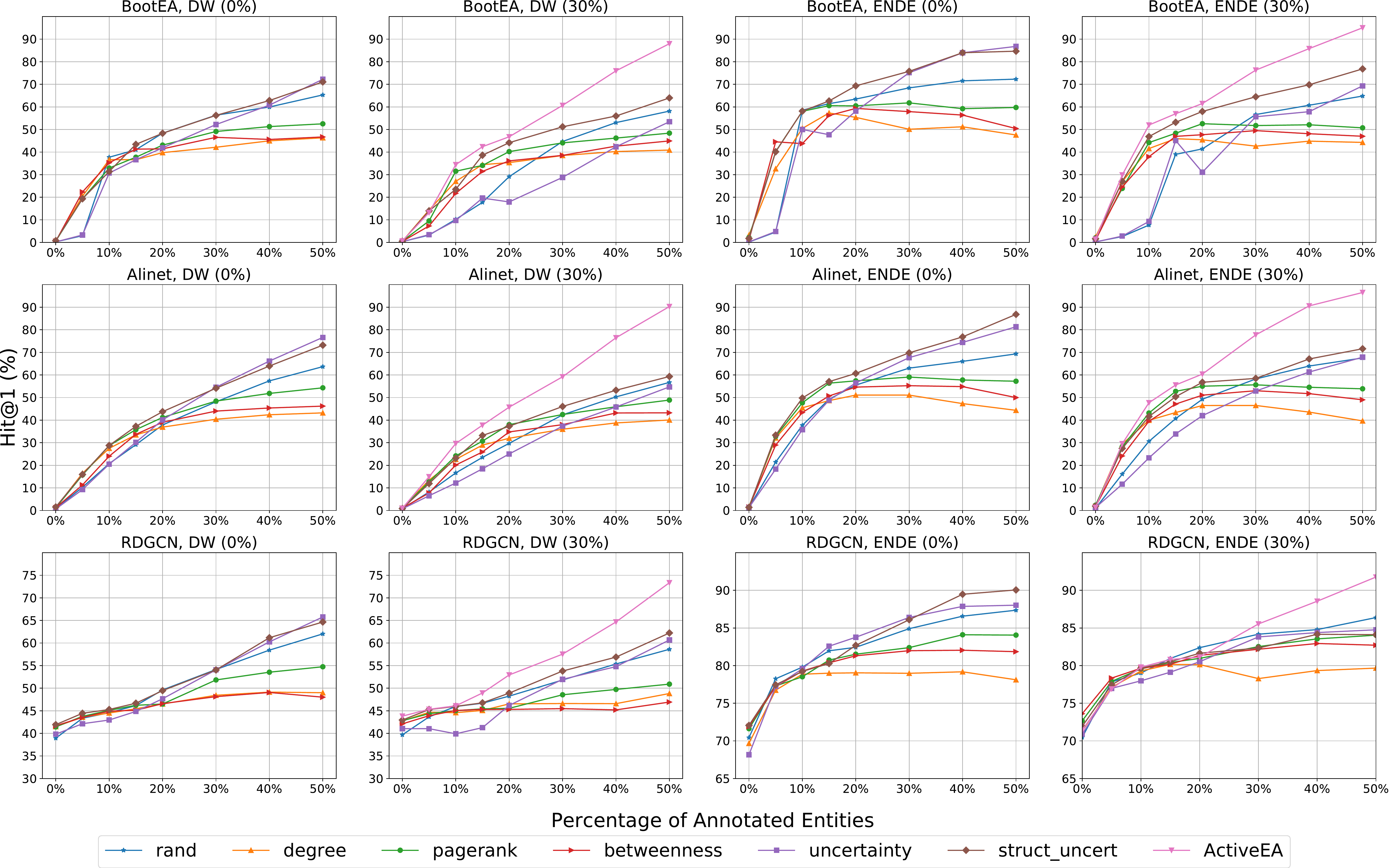}
	\caption{HIT@1 of sampling strategies for all EA models on DW and ENDE, as annotation portion increases. Top row shows experiments that do not include bachelors; bottom row shows experiments that include 30\% bachelors. \textit{ActiveEA} is equivalent to \textit{struct\_uncert} in absence of bachelors, and is thus shown only for the second row. \vspace{-12pt}}
	\label{fig:overall_performance}
\end{figure*}

\subsection{Reproducibility Details}

% Computing infrastructure:
Our experiments are run on a GPU cluster.
We allocate 50G memory and one 32GB nVidia Tesla V100 GPU for each job on 15K data, and 100G memory for each job on 100K data.
The training and evaluation of \textit{ActiveEA} take approximately 3h with Alinet on 15K data, 10h with BootEA on 15K data, 10h with RDGCN on 15K data, and 48h with Alinet on 100K data.
Most baseline strategies take less time than \textit{ActiveEA} on the same dataset except \textit{betweenness} on 100K data, which takes more than 48h.
We apply grid search for setting $\alpha$ and $N$ (shown in Sec.~\ref{sec:sensitivity_of_parameter}).
Hyper-parameters of the bachelor recognizer are chosen by referring the settings of OpenEA and our manual trials.
Code and datasets are available at \url{https://github.com/UQ-Neusoft-Health-Data-Science/ActiveEA}.

\begin{figure}[t!]
	\centering
	\includegraphics[width=\columnwidth]{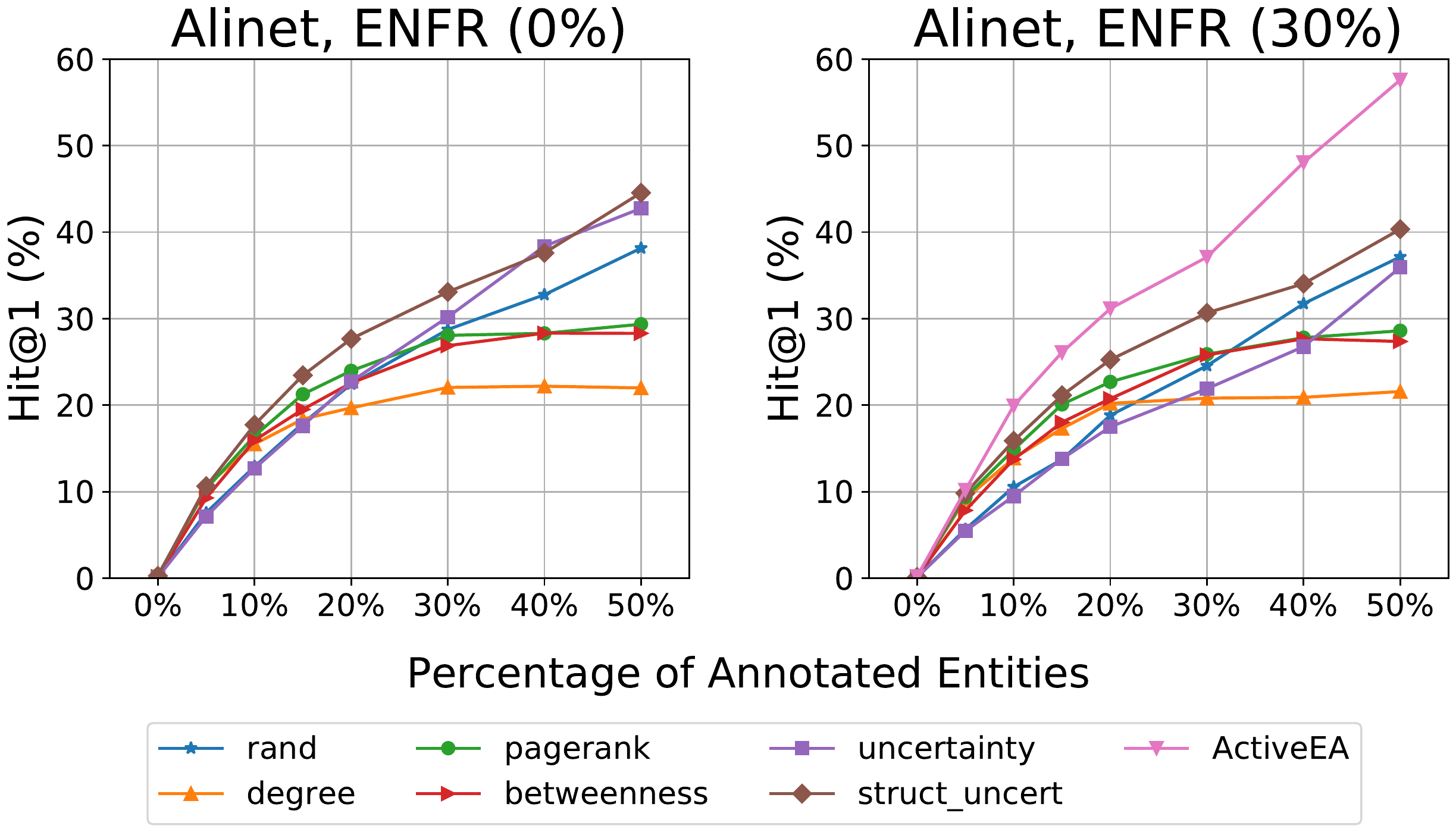}
	\caption{Hit@1 for all sampling strategies on the Alinet EA model on ENFR. Left shows experiments without bachelors, right shows with 30\% bachelors.}
	\label{fig:overall_performance_enfr100k}
\end{figure}

\begin{table*}
	\centering
	\scalebox{0.95}[0.95]{
\begin{tabular}{|l|c|c|c|c|c|c|c|c|c|c|c|c|}
        \hline
        \multirow{3}{*}{Strategy}
         & \multicolumn{4}{c|}{BootEA}  & \multicolumn{4}{c|}{AliNet} & \multicolumn{4}{c|}{RDGCN} \\
		\cline{2-13}
        %  & \scriptsize{DW(0\%)}  & \scriptsize{DW(30\%) }& \scriptsize{ENDE(0\%)} & \scriptsize{ENDE(30\%)}  & \scriptsize{DW(0\%)} & \scriptsize{DW(30\%)} & \scriptsize{ENDE(0\%)} & \scriptsize{ENDE(30\%)} & \scriptsize{DW(0\%)} & \scriptsize{DW(30\%)} & \scriptsize{ENDE(0\%)} & \scriptsize{ENDE(30\%)} \\
         & \scriptsize{\vtop{\hbox{\strut DW}\hbox{\strut (0\%)}}}   & \scriptsize{\vtop{\hbox{\strut DW}\hbox{\strut (30\%)}}}  & \scriptsize{\vtop{\hbox{\strut ENDE}\hbox{\strut (0\%)}}}  & \scriptsize{\vtop{\hbox{\strut ENDE}\hbox{\strut (30\%)}}}   & \scriptsize{\vtop{\hbox{\strut DW}\hbox{\strut (0\%)}}}  & \scriptsize{\vtop{\hbox{\strut DW}\hbox{\strut (30\%)}}}  & \scriptsize{\vtop{\hbox{\strut ENDE}\hbox{\strut (0\%)}}}  & \scriptsize{\vtop{\hbox{\strut ENDE}\hbox{\strut (30\%)}}}  & \scriptsize{\vtop{\hbox{\strut DW}\hbox{\strut (0\%)}}}  & \scriptsize{\vtop{\hbox{\strut DW}\hbox{\strut (30\%)}}}  & \scriptsize{\vtop{\hbox{\strut ENDE}\hbox{\strut (0\%)}}}  & \scriptsize{\vtop{\hbox{\strut ENDE}\hbox{\strut (30\%)}}}  \\
        \hline
        % \textit{rand} & 22.6 & 16.4 & 28.6 & 20.6 & 19.6 & 16.8 & 26.0 & 23.7 & 25.8 & 25.1 & 41.5 & 41.0 \\
        % \textit{degree} & 18.7 & 16.4 & 23.8 & 20.1 & 17.1 & 15.2 & 22.2 & 20.2 & 23.4 & 23.0 & 39.1 & 39.4 \\
        % \textit{pagerank} & 20.4 & 18.2 & 27.6 & 22.7 & 19.9 & 17.5 & 25.7 & 24.1 & 24.6 & 23.6 & 40.6 & 40.6 \\
        % \textit{betweenness} & 19.6 & 16.1 & 25.7 & 21.2 & 17.7 & 15.6 & 23.8 & 22.3 & 23.3 & 22.6 & 40.2 & 40.5 \\
        % \textit{uncertainty} & 21.6 & 12.8 & 29.7 & 20.0 & 21.9 & 14.8 & 27.7 & 21.4 & 25.8 & 24.3 & 41.8 & 40.6 \\
        % \hline
        % \textit{struct\_uncert} & 23.8 & 21.0 & 33.5 & 28.0 & 22.9 & 19.1 & 30.6 & 26.4 & 26.3 & 25.8 & 41.9 & 40.7 \\
        % \textit{ActiveEA} & - & 26.1 & - & 32.5 & - & 25.7 & - & 32.9 & - & 28.0 & - & 41.7 \\
        % \hline

        \hline
        rand & 23.5$^n$ & 17.0 & 28.1 & 21.3 & 19.4 & 16.7 & 26.0 & 23.7 & 25.8 & 25.0 & 41.3$^n$ & 41.0 \\
        degree & 19.5 & 16.0 & 24.0 & 20.0 & 17.1 & 15.2 & 22.2 & 20.5 & 23.3 & 22.9 & 39.1 & 39.4 \\
        pagerank & 22.3 & 18.3 & 27.6 & 23.0 & 19.9 & 17.3 & 25.8 & 24.1 & 24.5 & 23.9 & 40.5 & 40.6 \\
        betweenness & 20.5 & 16.3 & 26.1 & 21.1 & 17.8 & 15.6 & 23.7 & 22.3 & 23.2 & 22.7 & 40.2 & 40.3 \\
        uncertainty & 23.9 & 16.1 & 29.8 & 21.2 & 21.6 & 15.4 & 28.2 & 22.2 & 24.7 & 23.9 & 40.9$^n$ & 40.5 \\
        \hline
        struct\_uncert & \multirow{2}{*}{\textbf{26.3}}  & 20.8 & \multirow{2}{*}{\textbf{33.6}} & 27.4 & \multirow{2}{*}{\textbf{23.1}} & 19.1 & \multirow{2}{*}{\textbf{30.6}} & 26.8 & \multirow{2}{*}{\textbf{26.5}} & 25.6 & \multirow{2}{*}{\textbf{41.9}} & 41.0 \\
        ActiveEA & & \textbf{26.7} &  & \textbf{31.5} &  & \textbf{25.7} &  & \textbf{32.8} &  & \textbf{28.1} &  & \textbf{42.3} \\
        \hline

        % significance test - p values
        % rand & 0.01347 & 0.00000 & 0.00001 & 0.00108 & 0.00000 & 0.00000 & 0.00000 & 0.00000 & 0.00407 & 0.00017 & 0.02896 & 0.00136 \\
        % degree & 0.00011 & 0.00000 & 0.00000 & 0.00066 & 0.00000 & 0.00000 & 0.00000 & 0.00000 & 0.00001 & 0.00000 & 0.00007 & 0.00004 \\
        % pagerank & 0.00015 & 0.00000 & 0.00000 & 0.00200 & 0.00000 & 0.00000 & 0.00000 & 0.00000 & 0.00002 & 0.00002 & 0.00080 & 0.00035 \\
        % betweenness & 0.00014 & 0.00000 & 0.00000 & 0.00120 & 0.00000 & 0.00000 & 0.00000 & 0.00000 & 0.00000 & 0.00001 & 0.00037 & 0.00051 \\
        % uncertainty & 0.00033 & 0.00028 & 0.00015 & 0.00082 & 0.00031 & 0.00000 & 0.00042 & 0.00013 & 0.00637 & 0.00023 & 0.04232 & 0.00394 \\

        % significance test - corrected p value
        % rand & 0.06734 & 0.00000 & 0.00006 & 0.00539 & 0.00002 & 0.00000 & 0.00000 & 0.00001 & 0.02036 & 0.00085 & 0.14481 & 0.00681 \\
        % degree & 0.00053 & 0.00001 & 0.00000 & 0.00331 & 0.00000 & 0.00000 & 0.00000 & 0.00000 & 0.00003 & 0.00002 & 0.00036 & 0.00018 \\
        % pagerank & 0.00076 & 0.00001 & 0.00000 & 0.01001 & 0.00001 & 0.00000 & 0.00000 & 0.00001 & 0.00008 & 0.00009 & 0.00401 & 0.00174 \\
        % betweenness & 0.00068 & 0.00000 & 0.00000 & 0.00599 & 0.00000 & 0.00000 & 0.00000 & 0.00000 & 0.00002 & 0.00003 & 0.00185 & 0.00257 \\
        % uncertainty & 0.00167 & 0.00140 & 0.00074 & 0.00411 & 0.00153 & 0.00002 & 0.00210 & 0.00066 & 0.03186 & 0.00114 & 0.21161 & 0.01969 \\

\end{tabular}
}
	% \vspace{-8pt}
	\caption{Overall performance (AUC@0.5 (\%)) for each sampling strategy. The highest performing strategy in each column is indicated in bold. We run each strategy 5 times; most results for \textit{ActiveEA} show statistically significant differences over other methods (paired t-test with Bonferroni correction, $p<0.05$), except the few cells indicated by $^n$.}
	\label{tab:auc}
\end{table*}

\section{Experimental Results}

\subsection{Comparison with Baselines}\label{sec:overall_perf}
%\vspace{-4pt}
%\subsubsection{Overall Performance}

% We verify the effectiveness and generality of our AL strategies across different EA models, different datasets, and different proportions of bachelors. 
% We applied our methods and the baselines to three different EA models on two datasets, each of which we also synthetically modify to include 30\% bachelors.
% The curves in Fig.\ref{fig:overall_performance} show the effectiveness of our strategy and the baselines and each curve is about the performance of EA model with respect to the proportion of annotated entities.
Fig.~\ref{fig:overall_performance} presents the overall performance of each strategy with three EA models on two datasets, each of which we also synthetically modify to include 30\% bachelors. 
We also report the AUC@0.5 values of these curves in Tab.~\ref{tab:auc}.
\textit{ActiveEA} degenerates into \textit{struct\_uncert} when there is no bachelor.

% random sampling
\textbf{\textit{Random Sampling.}}
Random sampling usually performs poorly when the annotation proportion is small, while it becomes more competitive when the amount of annotations increases. But for most annotation proportions, random sampling exhibits a large gap in performance compared to the best method.
% The random sampling strategy achieves poor performance compared to other strategies. 
This observation highlights the need to investigate data selection for EA.

% findings about topology-based strategies
\textbf{\textit{Topology-based Strategies.}}
The topology-based strategies are effective when few annotations are provided, e.g., $<20\%$.  However, once annotations increase, the effectiveness of topology-based strategies is often worse than random sampling. This may be because these strategies suffer more from the bias between the training set and test set. Therefore, only considering the structural information of KGs has considerable drawbacks for EA. 
% This phenomenon reveals the necessity of considering the performance of EA model in the sampling process, as is done in the AL strategies of general domain.

% findings about uncertainty sampling
\textbf{\textit{Uncertainty Sampling.}}
On the contrary, the uncertainty sampling strategy performs poorly when the proportion of annotations is small but improves after several annotations have been accumulated. One reason for this is that neural EA models cannot learn useful patterns with a small number of annotations. On datasets with bachelors, uncertainty sampling always performs worse than random sampling. Thus, it is clear that uncertainty sampling cannot be applied directly to EA.

% findings about structure-aware uncertainty ssampling
%\noindent
\textbf{\textit{Structure-aware Uncertainty Sampling.}}
Struc\-ture-a\-ware uncertainty is effective across all annotation proportions. One reason for this is that it combines the advantages of both topology-based strategies and uncertainty sampling. This is essential for AL as it is impossible to predict the amount of annotations required for new datasets.

% findings about structure-awre uncertainty + bachelor recognizer
%\noindent
\textbf{\textit{ActiveEA.}}
\textit{ActiveEA}, which enhances structure-aware sampling with a bachelor recognizer, greatly improves EA when KGs contain bachelors.%brought great improvement when the KGs to align contained bachelor entities.

\vspace{-4pt}
\subsubsection{Generality}
\vspace{-2pt}
The structure-aware uncertainty sampling mostly outperforms the baselines, while \textit{ActiveEA} performs even better in almost all cases. \textit{ActiveEA} also demonstrates generality across datasets, EA models, and bachelor proportions.

% structure aware uncertainty vs. uncertainty 
%\textbf{\textit{Structure-aware Uncertainty vs. Uncertainty.}}
When the dataset has no bachelors, our uncertainty-aware sampling is exceeded by uncertainty sampling in few large-budget cases. However, the real-world datasets always have bachelors. 
In this case, our structure-aware uncertainty shows more obvious advantages.% However, this rarely happen when the datasets contains bachelors, which is much more common in real datasets.

% structure-aware uncertaint on RDGCN
%\textbf{\textit{Structure-aware Uncertainty and RDGCN.}}
In addition, the strategies are less distinguishable when applied to RDGCN. The reason is that RDGCN exploits the name of entities for pre-alignment and thus all strategies achieve good performance from the start.

% on large dataset
%\textbf{\textit{Experiments on Large Datasets.}}
To assess the generality across datasets of different sizes, we evaluate the sampling strategies with Alinet using {ENFR} (100K entities), which is larger than {DW} and {ENDE} (15K entities). We choose Alinet because it is more scalable than BootEA and RDGCN~\cite{zhao2020experimental}. %, which were also shown bad scalability in one recent existing survey work~\cite{zhao2020experimental}. 
Fig.~\ref{fig:overall_performance_enfr100k} presents comparable results to the 15K datasets.

\subsection{Effect of Bachelors}

\begin{figure}
	% \vspace{-8pt}
    \centering
    \includegraphics[width=\columnwidth]{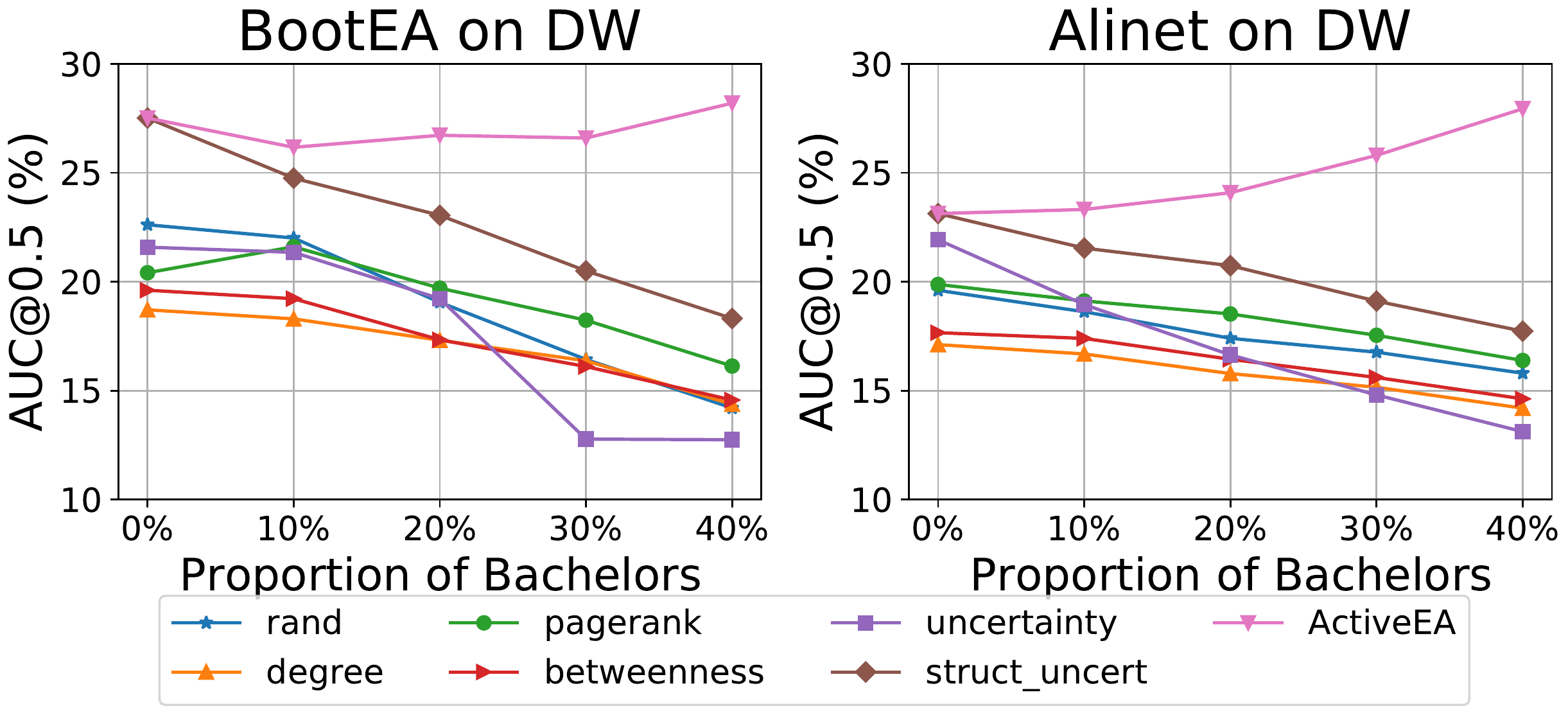}
    \caption{Comparison demonstrating the effect of bachelors (0\%~--~40\%) on the BootEA and Alinet models.}
    \label{fig:effect_of_bachelor}
\end{figure}

\begin{figure}
	\includegraphics[width=\columnwidth]{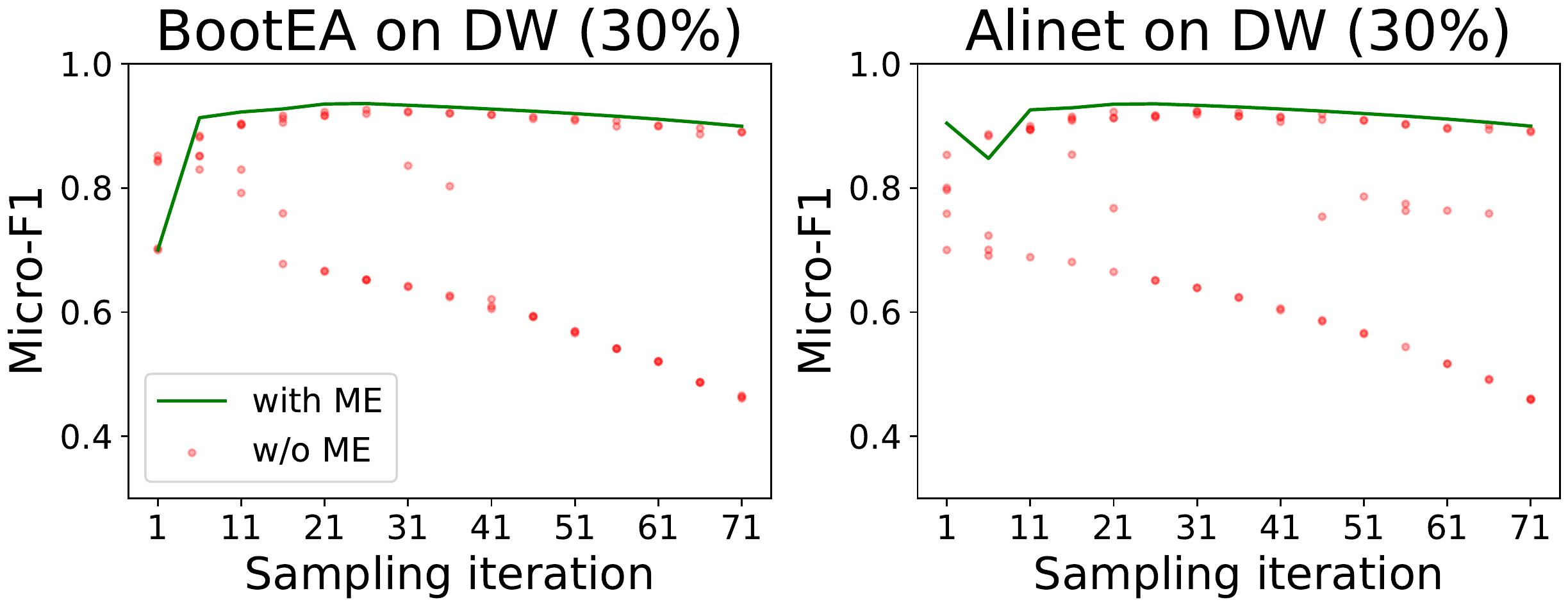}
	\caption{Comparison demonstrating the effectiveness of the bachelor recognizer and the effect of the model ensemble (ME) on BootEA and Alinet.}
	\label{fig:effectiveness_of_bachelor_recog}
\end{figure}

To investigate the effect of bachelors, we removed different amounts of entities randomly (each larger sample contains the subset from earlier samples) from $\mathcal{G}^2$ so that $\mathcal{G}^1$ had different percentages of bachelors.
Fig.~\ref{fig:effect_of_bachelor} shows the results of applying all strategies to these datasets. 
We further make the following four observations:
\begin{enumerate}[nolistsep,noitemsep,leftmargin=*,wide=0pt]
\item The performance of all strategies except \textit{ActiveEA} decrease as bachelors increase. How to avoid selecting bachelors is an important issue in designing AL strategies for EA.
\item Among all strategies, uncertainty sampling is affected the most, while topology-based methods are only marginally affected.
\item Our structure-aware uncertainty outperforms the baselines in all tested bachelor proportions.
\item \textit{ActiveEA} increases performance as the proportion of bachelors increases. The reason is: if $\mathcal{G}^1$ is fixed and the bachelors can be recognized successfully, a certain budget can lead to larger ratio of annotated matchable entities in datasets with more bachelors than in those with less bachelors.
% that the proportion of matchable entities increases as more bachelors are successfully removed.% the proportion of annotated matchable entities in the dataset with more bachelors was larger if the bachelors could be successfully removed.
\end{enumerate}

\subsection{Effectiveness of Bachelor Recognizer}

Fig.~\ref{fig:effectiveness_of_bachelor_recog} shows the effectiveness of our bachelor recognizer in the sampling process and the effect of model ensemble.
The green curve shows the Micro-F1 score of our bachelor recognizer using the model ensemble. Our bachelor recognizer achieves high effectiveness from the start of sampling, where there are few annotations.
Each red dot represents the performance of the bachelor recognizer trained with a certain data partition without using the model ensemble. Performance varied because of the bias problem. %There was a high likelihood that the bachelor recognizer achieved a poor performance. 
Therefore, our model ensemble makes the trained model obtain high and stable performance.

\begin{figure}[t!]
	\includegraphics[width=\columnwidth]{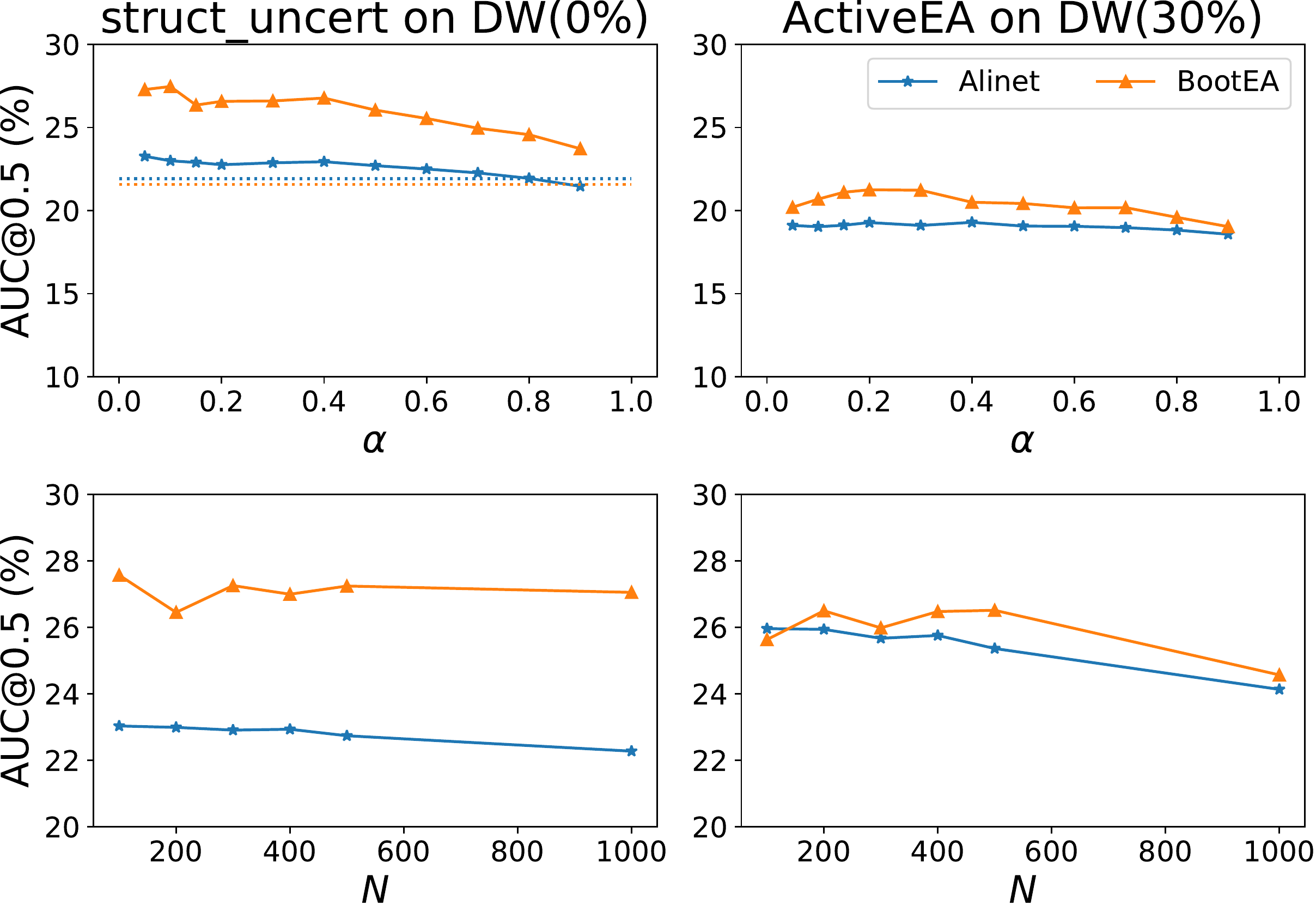}
	\caption{Comparison demonstrating the effects different parameters have on our sampling strategies.\vspace{-12pt}}
	\label{fig:sensitivity_of_param}
\end{figure}

\subsection{Sensitivity of Parameters}\label{sec:sensitivity_of_parameter}

% \subsubsection{Sensitivity of $\alpha$}

To investigate the sensitivity of parameters, we ran our strategy with AliNet and BootEA on two DW variants with bachelor proportions of 0\% and 30\%.% We report the performance of structure-aware uncertainty sampling when there is no bachelor.

The sensitivity w.r.t. $\alpha$ is shown in the top row of Fig.~\ref{fig:sensitivity_of_param}. We observe that our method is not sensitive to $\alpha$. The effectiveness fluctuates when $\alpha<0.5$, and decreases when $\alpha>0.5$. This indicates uncertainty is more informative than structural information. 
When $\alpha=0$, our struct\_uncert degenerates to uncertainty sampling (Eq.~\ref{eq:uncertainty}). In the upper left plot, we show the corresponding performance with dotted lines. Under most settings of $\alpha$, the struct\_uncert is much better than uncertainty sampling. This means that introducing structure information is beneficial.

% Therefore, we prefer to set $\alpha$ as a certain value less than 0.5. 
% \subsubsection{Effect of Sampling Batch Size}

The bottom row of Fig.~\ref{fig:sensitivity_of_param} shows the effect of sampling batch size $N$. The overall trend is that larger batch sizes decrease performance. This observation confirms the intuition that more frequent updates to the EA model lead to more precise uncertainty.
Therefore, the choice of value of sampling batch size is a matter of trade-off between computation cost and sampling quality.
%\todo{TODO: one more run to make the curves smoother}

% \begin{figure}[t]
%     \includegraphics[width=8.5cm]{images/effect_of_batchsize.png}
%     \caption{Effect of sampling batch size.}
%     \label{fig:effect_of_batchsize}
% \end{figure}

\subsection{Examination of Bayesian Transformation}

\begin{figure}[t]
    \includegraphics[width=\columnwidth]{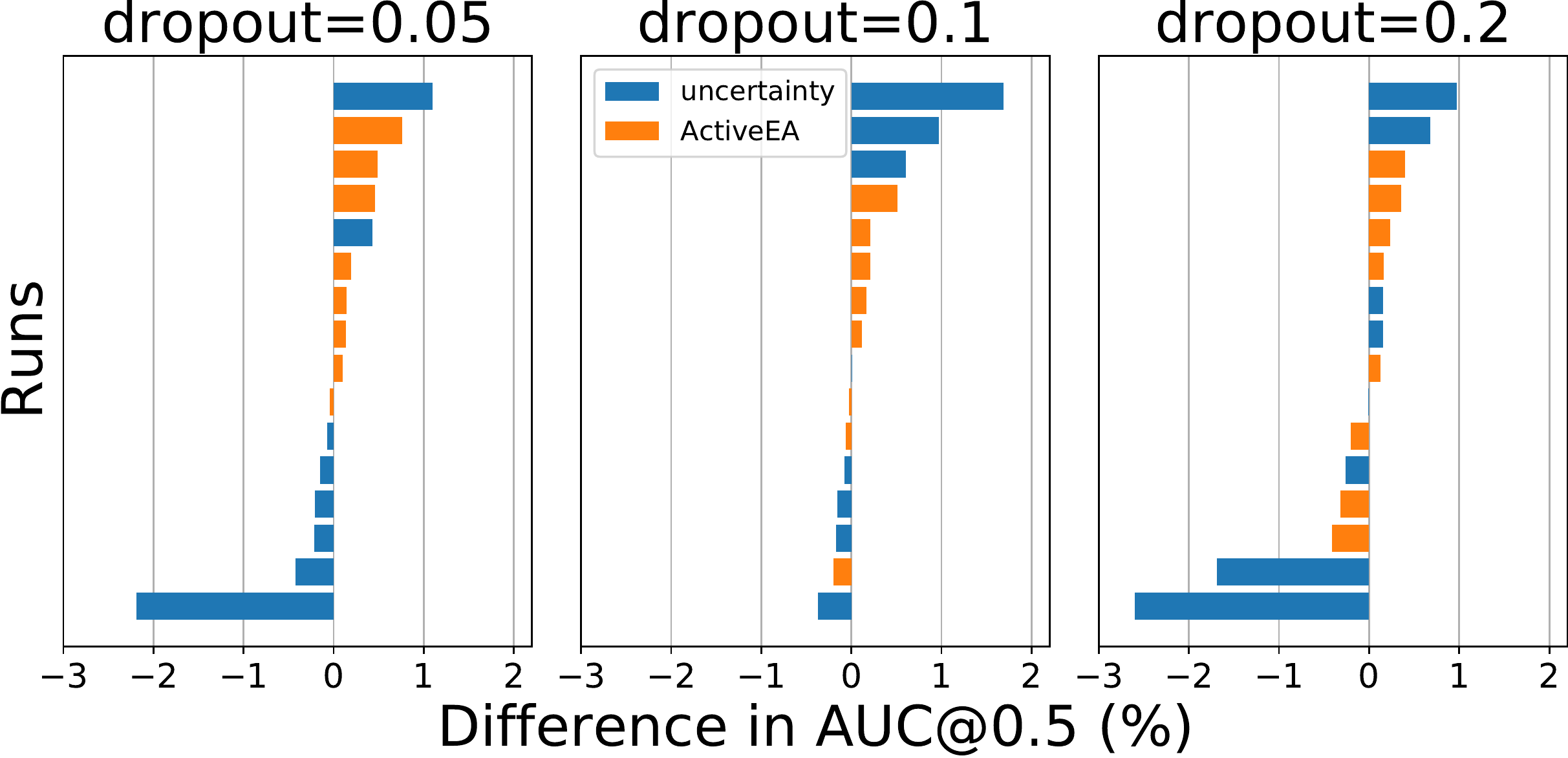}
    \caption{Effect of Bayesian Transformation on \textit{uncertainty} and \textit{ActiveEA} across the DW and ENDE datasets and different bachelor percentages.\vspace{-16pt}}
    \label{fig:effect_of_bayesian}
\end{figure}

We enhanced the uncertainty sampling and \textit{ActiveEA} with Bayesian Transformation, implemented with Monte Carlo (MC) dropout, and applied them to Alinet and RDGCN on DW and ENDE as in Sec.~\ref{sec:overall_perf}.
Fig.~\ref{fig:effect_of_bayesian} shows improvements with different settings of MC dropout rate. We find (1) the variation of effects on \textit{uncertainty} sampling is greater than that on \textit{ActiveEA}; (2) Bayesian Transformation with small dropout (e.g., 0.05) results in slight improvements to \textit{ActiveEA} in most cases.

\section{Related Works}

%\subsection{Entity Alignment}
\textbf{Entity Alignment}.~
Entity Alignment refers to the matching of entities across different KGs that refer to the same real-world object.
Compared with Entity Resolution~\cite{DBLP:conf/sigmod/MudgalLRDPKDAR18}, which matches duplicate entities in relational data, EA deals with graph data and emphasizes on exploiting the structure of KGs.
Neural models~\cite{DBLP:conf/ijcai/ChenTYZ17,DBLP:conf/ijcai/ChenTCSZ18,DBLP:conf/emnlp/WangLLZ18,DBLP:conf/acl/CaoLLLLC19} replaced conventional approaches~\cite{DBLP:conf/semweb/Jimenez-RuizG11,DBLP:journals/pvldb/SuchanekAS11} as the core methods used in recent years. 
Typically they rely on seed alignment as training data -- this is expensive to annotate.
Iterative training (i.e., self-training) has been applied to improve EA models by generating more training data automatically~\cite{DBLP:conf/ijcai/SunHZQ18,DBLP:conf/wsdm/MaoWXLW20}.
These works concern better training methods with given annotated data.
However, the problem of reducing the cost of annotation has been neglected.
\citet{DBLP:conf/ecir/BerrendorfFT21} have been the first to explore AL strategies for EA task. They compared several types of AL heuristics including node centrality, uncertainty, graph coverage, unmatchable entities, etc. and they empirically showed the impact of sampling strategies on the creation of seed alignment. In our work, we highlight the limitations of single heuristics and propose an AL framework that can consider information structure, uncertainty sampling and unmatchable entities at the same time.
% \citet{DBLP:conf/ecir/BerrendorfFT21} empirically showed the impact of sampling strategies on the creation of seed alignment but did not propose a specific solution for EA. Our work aims to fill this gap.
In addition, existing neural models assume all KGs entities have counterparts: this is a very strong assumption in reality~\cite{zhao2020experimental}.
We provide a solution to recognizing the bachelor entities, which is complementary to the existing models.

\noindent
\textbf{Active Learning}.~
Active Learning is a general framework for selecting the most informative data to annotate when training Machine Learning models~\cite{DBLP:books/crc/aggarwal14/AggarwalKGHY14}. 
% It's goal is to select the most informative data instances to label so as to get better models and save annotation budget.
The pool-based sampling scenario is a popular AL setting where a base pool of unlabelled instances is available to query from~\cite{DBLP:series/synthesis/2012Settles,DBLP:books/crc/aggarwal14/AggarwalKGHY14}.
Our proposed AL framework follows this scenario.
Numerous AL strategies have been proposed in the general domain~\cite{DBLP:books/crc/aggarwal14/AggarwalKGHY14}. \textit{Uncertainty sampling} is the most widely used because of its ease to implement and its robust effectiveness~\cite{DBLP:journals/sigir/Lewis95a,DBLP:journals/jair/CohnGJ96}.
However, there are key challenges that general AL strategies cannot solve when applying AL to EA. 
Most AL strategies are designed under the assumption that the data is independent and identically distributed. However, KGs entities in the AL task are correlated, as in other graph-based tasks, e.g., node classification~\cite{DBLP:conf/icml/BilgicMG10} and link prediction~\cite{DBLP:conf/www/OstapukYC19}. 
In addition, bachelor entities cause a very special issue in EA. They may have low informativeness but high uncertainty.
We design an AL strategy to solve these special challenges.
Few existing works~\cite{DBLP:conf/cikm/QianPS17,DBLP:conf/cikm/MalmiGT17} have applied AL to conventional EA but do not consider neural EA models, which have now become of widespread use.
Only \revised{\citet{DBLP:conf/ecir/BerrendorfFT21}} % Berrendorf et al.~\cite{DBLP:conf/ecir/BerrendorfFT21} 
empirically explored general AL strategies for neural EA but did not solve the aforementioned challenges.

\section{Conclusion}

Entity Alignment is an essential step for KG fusion. Current mainstream methods for EA are neural models, which rely on seed alignment.
The cost of labelling seed alignment is often high, but how to reduce this cost has been neglected.
In this work, we proposed an Active Learning framework (named ActiveEA), aiming to produce the best EA model with the least annotation cost.
Specifically, we attempted to solve two key challenges affecting EA that general AL strategies cannot deal with.
Firstly, we proposed a structure-aware uncertainty sampling, which can combine uncertainty sampling with the structure information of KGs.
Secondly, we designed a bachelor recognizer, which reduces annotation budget by avoiding the selection of bachelors. Specially, it can tolerate  sampling biases. %and it can tolerate  sampling biases. 
Extensive experimental showed ActiveEA is more effective than the considered baselines and has great generality across different datasets, EA models and bachelor percentages.
% \url{https://bit.ly/3fjNLuI}.

In future, we plan to explore combining active learning and self-training which we believe are complementary approaches. Self-training can generate extra training data automatically but suffers from incorrectly labelled data. This can be addressed by amending incorrectly labelled data using AL strategies.

\section*{Acknowledgements}
This research is supported by the Shenyang Science and Technology Plan Fund (No. 20-201-4-10), the Member Program of Neusoft Research of Intelligent Healthcare Technology, Co. Ltd.(No. NRMP001901)). Dr Wen Hua is the recipient of an Australian Research Council DECRA Research Fellowship (DE210100160). Dr Guido Zuccon is the recipient of an Australian Research Council DECRA Research Fellowship (DE180101579).

\bibliographystyle{acl_natbib}
\bibliography{al4ea}

\end{document}